\documentclass[conference]{IEEEtran}

\IEEEoverridecommandlockouts
\usepackage{cite}
\usepackage{amsmath,amssymb,amsfonts}
\usepackage{algorithmic}
\usepackage{algorithm}
\usepackage{float}
\usepackage{graphicx}
\usepackage{textcomp}
\usepackage{bm}
\usepackage{amsmath}
\usepackage{dblfloatfix}
\usepackage[bottom]{footmisc}
\usepackage{xcolor}
\usepackage{hyperref}
\def\BibTeX{{\rm B\kern-.05em{\sc i\kern-.025em b}\kern-.08em
    T\kern-.1667em\lower.7ex\hbox{E}\kern-.125emX}}
\usepackage{bm}

\title{Label-dependent and event-guided interpretable disease risk prediction using EHRs\\
}

\makeatletter

\author{
  \IEEEauthorblockN{Shuai NIU\textsuperscript{$^1$}, Yunya SONG\textsuperscript{$^2$},  Qing YIN\textsuperscript{$^1$}, Yike GUO\textsuperscript{$^1$}, Xian YANG\textsuperscript{$^{1,\ast}$ \thanks{* This is the corresponding author.}}}
  \IEEEauthorblockA{
  \textit{\textsuperscript{$^1$}The Department of Computer Science} \\
  \textit{\textsuperscript{$^2$}The Department of Journalism} \\
  \textit{Hong Kong Baptist University, Hong Kong, China}\\
    20483007@life.hkbu.edu.hk,yunyasong@hkbu.edu.hk, 21481326@life.hkbu.edu.hk \{yikeguo, xianyang\}@hkbu.edu.hk}
}

\begin{document}

\maketitle

\begin{abstract}
Electronic health records (EHRs) contain patients' heterogeneous data that are collected from medical providers involved in the patient’s care, including medical notes, clinical events, laboratory test results, symptoms, and diagnoses. In the field of modern healthcare, predicting whether patients would experience any risks based on their EHRs has emerged as a promising research area, in which artificial intelligence (AI) plays a key role. To make AI models practically applicable, it is required that the prediction results should be both accurate and interpretable. To achieve this goal, this paper proposed a label-dependent and event-guided risk prediction model (LERP) to predict the presence of multiple disease risks by mainly extracting information from unstructured medical notes. Our model is featured in the following aspects. First, we adopt a label-dependent mechanism that gives greater attention to words from medical notes that are semantically similar to the names of risk labels. Secondly, as the clinical events (e.g., treatments and drugs) can also indicate the health status of patients, our model utilizes the information from events and uses them to generate an event-guided representation of medical notes. Thirdly, both label-dependent and event-guided representations are integrated to make a robust prediction, in which the interpretability is enabled by the attention weights over words from medical notes. To demonstrate the applicability of the proposed method, we apply it to the MIMIC-III dataset, which contains real-world EHRs collected from hospitals. Our method is evaluated in both quantitative and qualitative ways.

\end{abstract}

\begin{IEEEkeywords}
Label-dependent prediction, Event-guided prediction, Cross-attention mechanism, Disease risk prediction. 
\end{IEEEkeywords}

\section{Introduction}
Artificial intelligence (AI) is being increasingly applied to extract information from electronic health records (EHRs) for implementing various prediction tasks, such as ICU staying time estimation \cite{xu2018raim }, disease diagnosis \cite{harutyunyan2019multitask,mullenbach2018explainable}, statistical phenotype prediction\cite{harutyunyan2019multitask}, and etc. 
EHRs collect heterogeneous information about the patients  from medical providers involved in the patients' care, including medical notes, laboratory observations, treatments, clinical events, electrocardiogram waveforms (ECG), and medication. 

 With rapid advances in deep learning, many methods like attention-based RNN \cite{xu2018raim} and convolutional neural networks (CNN) \cite{che2017exploiting} are being developed to predict disease risks using EHRs. To make these models practically useful, the predictive model is required to generate interpretable results while still retaining predictive power. However, for rare diseases, aforementioned approaches would not be applicable due to the lack of prior knowledge. 
 
 This paper aims to develop an AI model to fulfil the disease risk prediction task, and we are interested in utilizing attention-based methods to achieve interpretability for the task of risk prediction using medical notes. Our approach is different from self-attention-based methods\cite{vaswani2017attention} which did not use any external information to learn the important weights of words in medical notes. We propose a label-dependent and event-guided risk prediction (LERP) model that both names of disease risk labels and clinical events would be used to determine the importance of different words from medical notes. Apart from using the names of disease risk labels, we also use clinical events to set the attention weights of words from medical notes. Clinical events are treatments received from clinicians, and thus can be very informative in reflecting the patient health status. Our contributions can be summarized as follows:
\begin{itemize}
\item We propose a cross-attention mechanism to learn the attention weights of words in medical notes by measuring their semantic similarities with names of disease risk labels and clinical events.
\item To encode textual information, we apply a pre-trained biomedical language model, Clinical-BERT \cite{alsentzer2019publicly}, for jointly embedding names of disease risk labels, clinical events, and medical notes such that information learned from a large biomedical corpus can naturally be incorporated into the model. Label names, clinical events, and words in medical notes with similar meanings will be assigned with similar embedding vectors by Clinical-BERT.
\end{itemize}



\section{Related Work}
\subsection{Label-dependent Predictive Modelling}
Label-dependent predictive models are being developed in various domains, such as computer vision (CV) (e.g., object detection \cite{kamath2021mdetr}) and modern healthcare (e.g., disease codes prediction \cite{mullenbach2018explainable}).
 In the medical healthcare domain, \cite{mullenbach2018explainable} first proposed the convolutional attention for a multi-label classification  model (CAML) and deep CAML to predict multiple diseases by introducing the label information via a attention layer. Following the work of CAML,  \cite{wang2018joint} proposed the label-embedding attentive model (LEAM) to jointly learn  the embeddings of medical  notes and label names in the same latent space. 
\subsection{Using Clinical Events for Disease Prediction}
Clinical events recorded in  EHRs indicate treatments that were given to  patients based on their own clinical conditions. Many researchers attempted to use clinical events for predictive model construction. \cite{xu2018raim} treated clinical events as interventions and also adopted the attention mechanism to generate the weighted embedding of electrocardiogram (ECG)  for patients' mortality prediction. \cite{choi2017using} adopted the gated recurrent units (GRUs) to detect relationships among various time-stamped events for the heart failure prediction.

\section{Methods}
\subsection{ Problem Definition and Notations}
In this work, we focus on using medical notes and clinical events to predict whether patients would experience some disease risks. 
Let us first define the vocabulary of words that occurred across all EHRs as  $V$, whose size is represented as $|V|$.
The information from each EHR used for the risk prediction is defined as $\bm{X}=\{ \bm{M}, \mathcal{L}_E, \mathcal{L}_Y \}$.
Here,
$\bm{M} = \{\bm{m}_1,...,\bm{m}_{N_M}\}$ contains a sequence of words from a medical note; $\mathcal{L}_E =\{\bm{l}_1,...,\bm{l}_{N_E}\}$ refers to a set of clinical events; and $\mathcal{L}_Y =\{\bm{l}_1,...,\bm{l}_{N_Y}\}$ represents the names of disease risk labels.
Each element from $\bm{M}$, $\mathcal{L}_E$, $\mathcal{L}_Y$ is a $|V|$-dimensional one-hot vector for representing  a word, an event and the name of a risk label, respectively. 
Please note that for each EHR, $\bm{M}$ and $\mathcal{L}_E$ are different but $\mathcal{L}_Y$ is identical. This is because $\mathcal{L}_Y$  just encodes names of disease risk labels and does not indicate their presence.
A sample from the training dataset is represented as $(\bm{X}, \bm{y})$, where $\bm{y} \in \mathcal{Y}$ is a $N_Y$-dimensional vector with elements equal to 1 or 0 indicating the presence of different disease risks.
 The goal for disease risk prediction is to learn a mapping function $f: \mathcal{X} \rightarrow \mathcal{Y}$ by minimizing the prediction loss.

\begin{figure}[htbp]
\centerline{\includegraphics[scale=0.30]{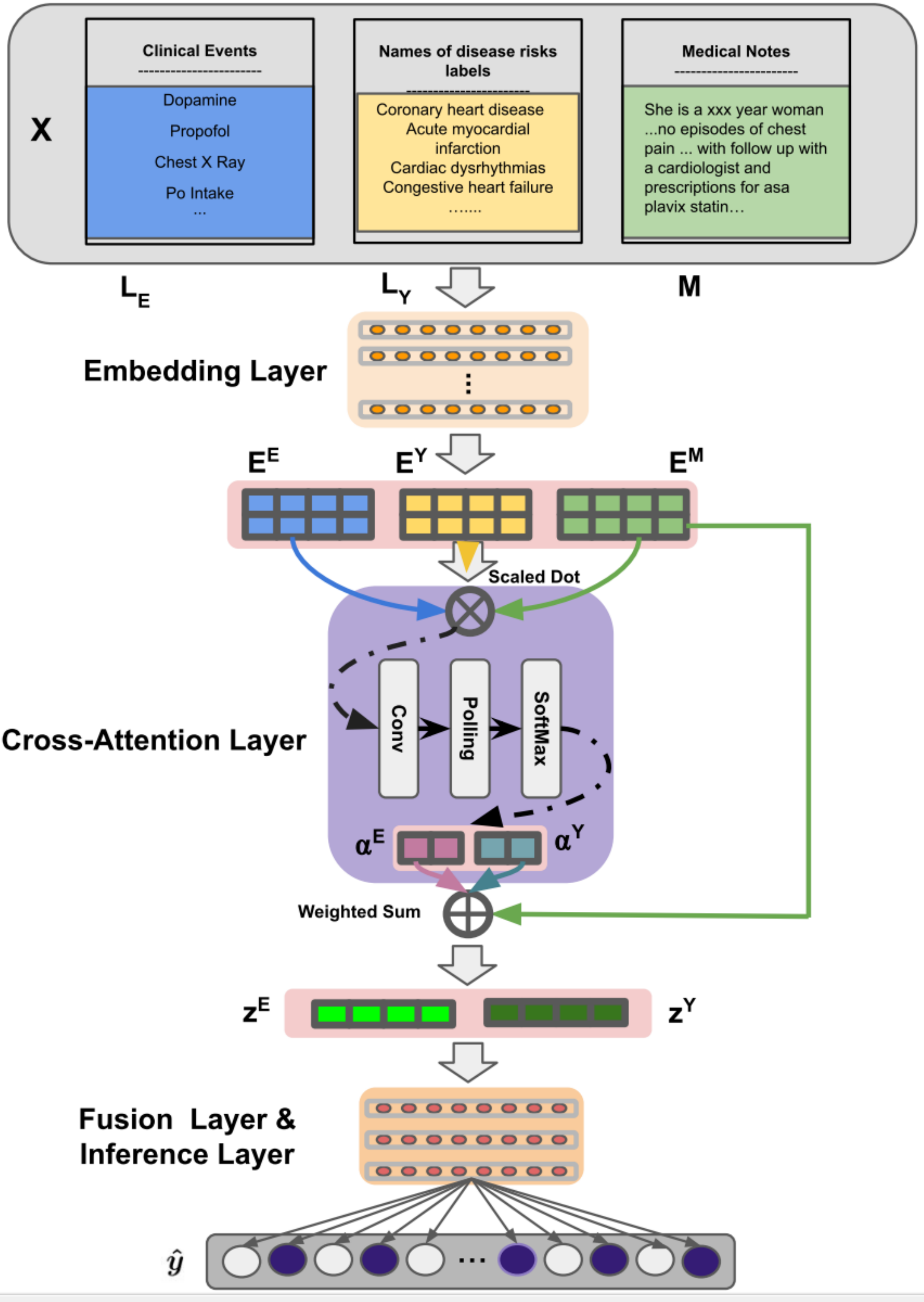}}
\caption{The structure of the LERP Model. It takes the information from medical notes, clinical events and names of disease risk labels as the inputs.  LERP is composed of embedding layers for textual information embedding, cross-attention layer for learning weighted representations of the medical note, and the fusion layer together with the output layer to predict the presence of different disease risks.}
\label{fig:model}
\end{figure}
\subsection{Model Overview}
Fig.~\ref{fig:model} shows the details of our proposed model, LERP. The text encoder based on Clinical-BERT first converts the medical note $\bm{M}$, the sequence of clinical events $\mathcal{L}_E$, and the names of disease risk labels $\mathcal{L}_Y$ into embedding matrices $\bm{E}^M$, $\bm{E}^E$, and $\bm{E}^Y$, respectively. 
Then the cross-attention mechanism is introduced to generate attention matrices  $\bm{\alpha}^E$ and $\bm{\alpha}^Y$. $\bm{\alpha}^E$ measures the similarities between elements from $\bm{E}^M$ and $\bm{E}^E$, while $\bm{\alpha}^Y$ is similarly calculated for $\bm{E}^Y$ and $\bm{E}^M$. With these two attention matrices, the model obtains two weighted representations of the medical note, denoted as  $\bm{z}^{E}$ and $\bm{z}^{Y}$. Our model uses the information encoded in  $\bm{z}^{E}$ and $\bm{z}^{Y}$ to predict the presence of $N_Y$ different disease risks. 

\subsection{Embedding Layer}
First, $\bm{M}$, $\mathcal{L}_E$, and $\mathcal{L}_Y$ are  
passed through an embedding layer $f_0$
to get $\bm{E}^M \in \mathbb{R}^{D \times N_M}$, $\bm{E}^E \in \mathbb{R}^{D \times N_E}$ and $\bm{E}^Y \in \mathbb{R}^{D \times N_Y}$,  where $D$ is the embedding size. In our model, $f_0$ is implemented by Clinical-BERT\cite{johnson2016mimic}. To get embedding matrices $\bm{E}^Y$ and $\bm{E}^E$, we use the averaged embeddings of input tokens to represent the overall embedding of an event or the name of a disease risk label. To generate $\bm{E}^M$, Clinical-BERT encodes medical notes and returns $N_M$ embedding vectors for all $N_M$ words.

\subsection{Cross-attention Layer}
The cross-attention layer is illustrated in the middle part of Fig.~\ref{fig:model}, where $\bm{E}^M$, $\bm{E}^E$, and  $\bm{E}^Y$ are the inputs.
We first apply a fully connected layer $f_1$ to reduce the embedding dimension of $\bm{E}^M$, $\bm{E}^E$, and  $\bm{E}^Y$ from  $D$ to $F$. 
The outputs of $f_1$ are then used to compute the scaled-dot similarity matrices $\bm{G}^E \in \mathbb{R}^{N_M \times N_E}$ and $\bm{G}^Y \in \mathbb{R}^{N_M \times N_Y}$:
\begin{equation}
\bm{G}^E = ScaledDot(f_1(\bm{E}^M), f_1(\bm{E}^E)) =  \frac{({f_1(\bm{E}^M)})^T* f_1(\bm{E}^E)}{\sqrt{F}}
\label{eq:1_1}
\end{equation}
where the $(.)^T$ is the transpose operator and $*$ is the matrix product operator. We use the same equation to calculate $\bm{G}^Y$ with the input of $\bm{E}^Y$ and $\bm{E}^M$. 

We use a one-dimensional (1-D) CNN with a max-pooling (MP) layer to better capture the relative spatial information of successive words and to increase the ability of implicit information extraction:
\begin{equation}
\bm{u}^E = 
MaxPool(ReLU(Conv(\bm{G}^E,k_1,q)), k_2)\label{eq:2}
\end{equation}
 where $ReLU$ is the nonlinear activation layer, $k_1$ is the kernel width (N-Gram) of CNN, $q$ is the padding size of CNN (set to `same padding' in our implementation), and $k_2$ is the kernel width of MP. The $\bm{u}^Y$ is generated by the same formula as $\bm{u}^E$ with input of $\bm{G}^Y$.

The outputs $\bm{u}^E \in \mathbb{R}^{N_M}$ and $\bm{u}^Y \in \mathbb{R}^{N_Y}$ are then normalized by a SoftMax function to generate $\alpha^E$ and $\alpha^Y$. With $\bm{\alpha}^E$ and $\bm{\alpha}^Y$, we can obtain the two weighted representations of the medical note as follows:
\begin{equation}
\bm{z}^E,\bm{z}^Y  = \sum_{n=1}^{N_M} \alpha_n^E \bm{E}_n^M,\sum_{n=1}^{N_M} \alpha_n^Y \bm{E}_n^M
\label{eq:5}
\end{equation}

where $\bm{E}_n^M \in \mathbb{R}^{D}$ is the $n$th column of $\bm{E}^M$, $\alpha_n^E$ and $\alpha_n^Y$ are the $n$th elements from $\bm{\alpha}^E$ and $\bm{\alpha}^Y$ respectively.


\subsection{Fusion and output Layers}
After we have obtained $\bm{z}^E$ and $\bm{z}^Y$, we combine them into one vector via fully connected layers $f_1$, $f_2$, and $f_3$:
\begin{equation}
\bm{\hat \bm{y}} = Sigmoid(f_3(f_1(f_2(\bm{z}^E \oplus \bm{z}^Y))),\label{eq:7}
\end{equation}
where $\oplus$ is the concatenation operator and $\bm{\hat \bm{y}} \in \mathbb{N}^{Y}$.

\subsection{Model Training}
To train our model, the loss for each EHR is defined as follows:
\begin{equation}
\begin{aligned}
Loss& = -\frac{1}{N_Y} \sum_{j=1}^{N_Y} (y_j \cdot \log(\hat y_j)) + (1-y_j)\cdot \log(1-\hat {y}_j)),
\label{eq:9}
\end{aligned}
\end{equation}
where $y_j \in \{0,1\}$ indicates the presence of the $j$th disease risk.

\section{Experiments}
\subsection{Experimental Dataset}
We evaluate the performance of our proposed LERP model on a public EHR dataset, MIMIC-III\cite{johnson2016mimic}.  In this paper, for disease risk prediction we only focus on using the information from medical notes of the discharge summary and clinical events. 
We choose  25 types of disease risks (defined in \cite{harutyunyan2019multitask}) as our prediction tasks, where some of them are clinically different. Across all EHRs, there are 1,152 distinct clinical events.
The MIMIC-III dataset contains 58,976 EHRs from 46,520 patients. We select 31,484 unique EHRs with no missing information.
The data pre-processing approach used in CAML is adopted to analyze the unstructured medical notes. For performance evaluation, we follow the data splitting strategy as used in \cite{harutyunyan2019multitask} to get 25,190 training and 6,294 testing samples (80\% for training and 20\% for validation).

\begin{table}[b]
\begin{center}

\caption{Performance of comparative methods}
\resizebox{230pt}{50pt}{

\begin{tabular}{|c|c|c|c|}
\hline
&\multicolumn{3}{|c|}{\textbf{Evaluation Metrics}} \\
\cline{2-4} 
\textbf{Models} & Micro Precision & Macro Precision & Micro Recall   \\

\hline
LEAM &  0.7526 &  0.6308 & 0.4958  \\
\hline
TS & 0.7256 & 0.6533 & 0.5968 \\
\hline

LERP & 0.7231 & 0.6645 & 0.6075   \\
\hline
LERP$\bm{^-}$ & 0.7075 & 0.6598 & 0.6305\\
\hline

\textbf{Models} &  Macro Recall & Micro ROC AUC &  Macro ROC AUC    \\
\hline
LEAM & 0.4347 & 0.8898 & 0.8587  \\
\hline
TS & 0.5404 & 0.8969 & 0.8642 \\
\hline

LERP &  0.5424 & 0.9001 & 0.8729  \\
\hline
LERP$\bm{^-}$ &  0.581 & 0.9013 & 0.8737\\
\hline
\end{tabular}
\label{tab1:performance}
}
\end{center}

\end{table}

\subsection{Comparative Methods and Implementation Details}
In order to make a comprehensive comparison, we compare our model with other comparative methods as described below:
\begin{itemize}
     \item \textbf{LEAM}: LEAM is a cutting-edge deep learning model that was created specifically for ICD-9 code prediction by utilizing textual information of medical notes. We select the default setting of LEAM as implemented in
    \footnote{\href{https://github.com/guoyinwang/LEAM}{https://github.com/guoyinwang/LEAM}} for comparison.
    \item \textbf{TS}: This baseline model applies Clinical-BERT\cite{huang2019clinicalBERT} to embed medical notes. The self-attention mechanism \cite{vaswani2017attention}
    is adopted to encode information from the medical notes for disease risk prediction.
  
   \item \textbf{LERP}:  Our LERP model\footnote{\href{https://github.com/finnickniu/LERP}{https://github.com/finnickniu/LERP}} is a label-dependent and event-guided approach to make
   interpretable risk predictions.  Medical notes, names of disease risk labels, and clinical events are embedded by Clinical-BERT. The cross-attention mechanism is introduced to assign attention weights to words from  medical notes based on the semantic similarities among words,  events, and names of disease risk labels.
  \item \textbf{LERP$\bm{^-}$}: This is a modified version of LERP that clinical events are not included in the risk prediction model.  Attentions of words from medical notes are determined by their semantic similarities with names of disease risk labels. 
\end{itemize}

\subsection{Quantitative analysis}

The performance of all comparative models is evaluated using the following metrics: precision, recall, and ROC AUC score. We compute both micro- and macro-averages for these metrics.  Table \ref{tab1:performance} shows the results of all comparative methods, from which we have the following observations:
\begin{itemize}
    \item Compared with LEAM which does not use Clinical-BERT for textual information embedding, our LERP model returns higher values for most evaluation metrics. 
    Especially, LEAM has much lower recall values. This is because Clinical-BERT can be useful in learning semantic representations of medical textual information.
    This observation demonstrates the power of incorporating the pretrained language model for the risk prediction. 
    \item Compared with TS which is not label-dependent, LERP returns higher values in most evaluation metrics as well.  This observation indicates that the cross-attention mechanism, making the predictive model label-dependent, would work better than the self-attention mechanism.
    \item Compared with LERP$\bm{^-}$ which does not use the information from clinical events, the values of evaluation metrics obtained from
    our full model are slightly lower but the difference in ROC AUC values is trivial. This is because LERP$\bm{^-}$ learns the attentions of words from medical notes fully dependent on the prediction tasks. In our full LERP model, attentions are also guided by clinical events. Although our full model has sacrificed a little bit of  performance, it would give better interpretable results which will be shown in the following subsection.

\end{itemize}

\begin{figure}[!b]
\centerline{\includegraphics[scale=0.38]{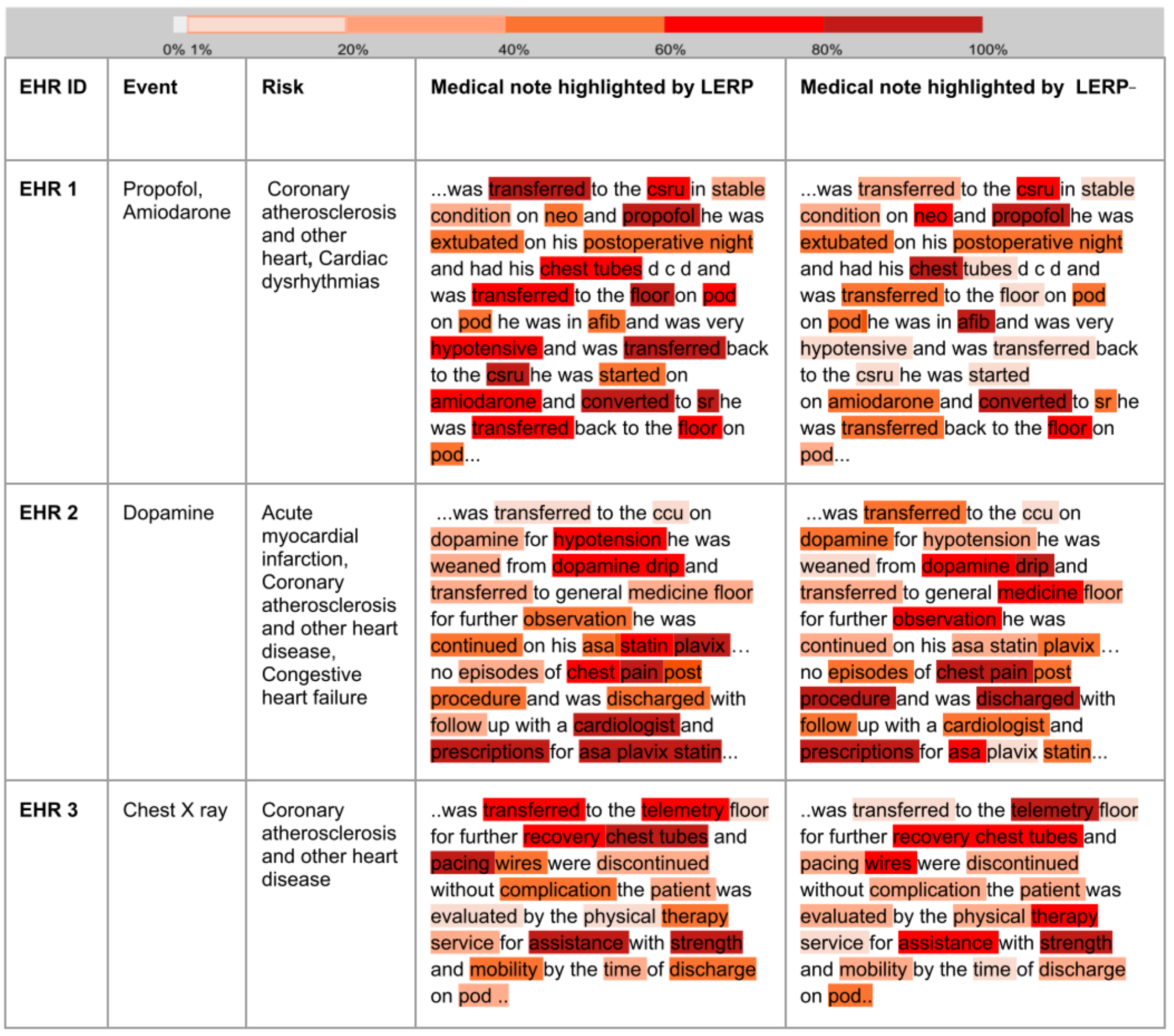}}
\caption{Case studies to compare the interpretable results from LERP and LERP$^-$. The colour map on the top of this figure maps the colours to normalized attention scores (ranging from 0\% to 100\%). In the result table, the second/third column contains the clinical events/disease risks associated with the selected fragments of medical notes.}
\label{fig:case_stduy}
\end{figure}

\subsection{Qualitative analysis}

In this subsection, we carried out case studies to show the interpretability of our model by
investigating which words from medical notes have gained high attention from our model and checking whether these words are clinically relevant to the risks. Three EHRs for patients with different disease risks have been randomly selected from the MIMIC-III dataset. Fig.~\ref{fig:case_stduy} shows fragments of medical notes, clinical events, and risks that have been recorded in each EHR. Words from EHR fragments are highlighted in red, whose darkness are determined by their attention scores derived from the cross-attention mechanism. Clinical events and disease risks that are associated with the given medical note fragments are given as well.
To show whether the event-guided approach would improve  interpretability, we compare results from LERP with LERP$\bm{^-}$.

The patient recorded in `EHR 1' has the risks of `Coronary atherosclerosis ...' and `Cardiac dysrhythmias'. By comparing the results from LERP$\bm{^-}$ and LERP, we can find that LERP, for example, gives higher attentions to the following two words from medical notes: `Amiodarone' and `hypotensive'.  `Amiodarone' is a medicine frequently used to treat  both `Coronary atherosclerosis ...' and `Cardiac dysrhythmias' (\cite{siddoway2003amiodarone}), while `hypotensive' is a typical symptom of these risks \cite{owens1999hypotension}. For the rest cases, we can also find similar result that LERP can capture more related clinical phases than LERP$\bm{^-}$.

\section{Conclusions}
This study presents an interpretable label-dependent and event-guided prediction model to predict the presence of various disease risks by using the names of disease risks, clinical events, and medical notes from EHRs. We employ Clinical-BERT as an embedding layer to assist our LERP model in extracting information from raw textual data. With the adoption of the cross-attention mechanism, representations of medical notes are generated by learning attention influenced by both clinical events and names of disease risk labels.
We evaluate our model LERP using the MIMIC-III dataset to show its predictive power and interpretability. 
Case studies have been conducted to show that
the medical terms that are clinically relevant to the disease risks gain high attention weights. In the future, we will invite domain experts to manually annotate our results, for example, to specify which words from medical notes are relevant to risk labels. As such, we can quantitatively evaluate the degree of interpretability.


{\small

\bibliographystyle{IEEEtranS}
}
\end{document}